\documentclass[11pt]{article}
\usepackage{coling2016}
\usepackage{times}
\usepackage{url}
\usepackage{graphicx, wrapfig}
\usepackage{latexsym}
\usepackage{amsfonts, amssymb, amsmath}
\usepackage{algorithm, algorithmic, amsthm}
\usepackage{graphicx}
\usepackage{color}
\makeatletter
\newcommand{\@BIBLABEL}{\@emptybiblabel}
\newcommand{\@emptybiblabel}[1]{}
\makeatother
\usepackage{hyperref}
\usepackage[top=2.5cm, bottom=3cm, left=3cm, right=3cm]{geometry}

\usepackage{subfigure}

{%
\begin{list}{#1}{
\vspace{-\topsep}
\vspace{-\partopsep}
\setlength{\itemindent}{0cm}
\setlength{\rightmargin}{0cm}
\setlength{\listparindent}{0cm}
\settowidth{\labelwidth}{#1}
\setlength{\leftmargin}{\labelwidth}
\addtolength{\leftmargin}{\labelsep}
\setlength{\itemsep}{0cm}
}%
}%
{%
\end{list}
\vspace{-\topsep}
\vspace{-\partopsep}
}

{\begin{enumerate}%
}%
{\end{enumerate}}

\theoremstyle{plain}

\newtheorem*{lemma*}{Lemma}

\newtheorem*{prop*}{Proposition}

\theoremstyle{definition}

\newtheorem*{defn*}{Definition}

\newtheorem*{exmp*}{Example}

\newtheorem*{conj*}{Conjecture}

\theoremstyle{remark}

\newtheorem*{rmk*}{Remark}

\title{Interactive Attention for Neural Machine Translation}

\author{Fandong Meng$^1$\thanks{\ \ The majority of this work was completed when the first author studied at Institute of Computing Technology, Chinese Academy of Sciences.}  \  \  Zhengdong Lu$^2$ \ Hang Li$^2$  \ Qun Liu$^{3,4}$ \\\\
        $^1$AI Platform Department, Tencent Technology Co., Ltd.\\
        {\tt fandongmeng@tencent.com}\\
        $^2$Noah's Ark Lab, Huawei Technologies\\
        {\tt \{Lu.Zhengdong,HangLi.HL\}@huawei.com}\\
        $^3$ADAPT Centre, School of Computing, Dublin City University\\
        $^4$Key Laboratory of Intelligent Information Processing, Institute of Computing Technology, CAS \\
        {\tt qliu@computing.dcu.ie}\\
}

\date{}

\begin{document}

\maketitle
\begin{abstract}
Conventional attention-based Neural Machine Translation (NMT) conducts dynamic alignment in generating the target sentence. By repeatedly reading the representation of source sentence, which keeps fixed after generated by the encoder~\cite{cho}, the attention mechanism has greatly enhanced state-of-the-art NMT. In this paper, we propose a new attention mechanism, called~\textsc{Interactive Attention}, which models the interaction between the decoder and the representation of source sentence during translation by both reading and writing operations. \textsc{Interactive Attention} can keep track of the interaction history and therefore improve the translation performance. Experiments on NIST Chinese-English translation task show that \textsc{Interactive Attention} can achieve significant improvements over both the previous attention-based NMT baseline and some state-of-the-art variants of attention-based NMT (i.e., coverage models~\cite{Tu2016}). And neural machine translator with our \textsc{Interactive Attention} can outperform the open source attention-based NMT system Groundhog by 4.22 BLEU points and the open source phrase-based system Moses by 3.94 BLEU points averagely on multiple test sets.
\end{abstract}

\section{Introduction}\label{intro}

\blfootnote{
    \hspace{-0.65cm}  
    This work is licensed under a Creative Commons Attribution 4.0 International Licence. Licence details: http://creativecommons.org/licenses/by/4.0/
}

Neural Machine Translation (NMT) has made promising progress in recent years~\cite{googleS2S,cho,luongEMNLP2015,jean2015using,luongEtAl2015,TangMLLY16,WangLLL16,litowards2016,Tu2016,ShenCHHWSL15,ZhouCWLX16}, in which attention model plays an increasingly important role. Attention-based NMT represents the source sentence as a sequence of vectors after a RNN or bi-directional RNN~\cite{schuster1997bidirectional},  and then simultaneously conducts dynamic alignment with a gating neural network and generation of the target sentence with another RNN. Usually NMT with attention model is more efficient than its attention-free counterpart: it can achieve comparable results with far less parameters and training instances~\cite{jean2015using}. This superiority in efficiency comes mainly from the mechanism of dynamic alignment, which avoids the need to represent the entire source sentence with a fixed-length vector~\cite{googleS2S}.

However, conventional attention model is conducted on the representation of source sentence (fixed after generated) only with reading operation~\cite{cho,luongEMNLP2015}. This may let the decoder tend to ignore past attention information, and lead to over-translation and under-translation~\cite{Tu2016}. To address this problem,~\newcite{Tu2016} proposed to maintain tag vectors in source representation to keep track of the attention history, which encourages the attention-based NMT system to consider more  untranslated source words. Inspired by neural turing machines~\cite{ntmgraves2014neural}, we propose \textsc{Interactive Attention} model from the perspective of memory reading-writing, which provides a conceptually simpler and practically more effective mechanism for attention-based NMT. The NMT with \textsc{Interactive Attention} is called NMT$_\textsf{IA}$, which can keep track of the interaction history with the representation of source sentence by both reading and writing operations during translation. This interactive mechanism may be helpful for the decoder to automatically distinguish which parts have been translated and which parts are under-translated.

We test the efficacy of NMT$_\textsf{IA}$ on NIST Chinese-English translation task. Experiment results show that NMT$_\textsf{IA}$  can significantly outperform both the conventional attention-based NMT baseline~\cite{cho} and coverage models~\cite{Tu2016}. And neural machine translator with our \textsc{Interactive Attention} can outperform the open source attention-based NMT system Groundhog by 4.22 BLEU points and the open source phrase-based system Moses by 3.94 BLEU points.

\paragraph{RoadMap:} In the remainder of this paper, we will start with a brief overview of attention-based neural machine translation in Section~\ref{back}. Then in Section~\ref{interactive}, we will detail the  \textsc{Interactive Attention}-based NMT (NMT$_\textsf{IA}$). In Section~\ref{experiments}, we report our empirical study of NMT$_\textsf{IA}$ on a Chinese-English translation task, followed by Section~\ref{related}
and~\ref{conclusion} for related work and conclusion.

\begin{figure}[t!]
\begin{center}
      \includegraphics[width=0.75\textwidth]{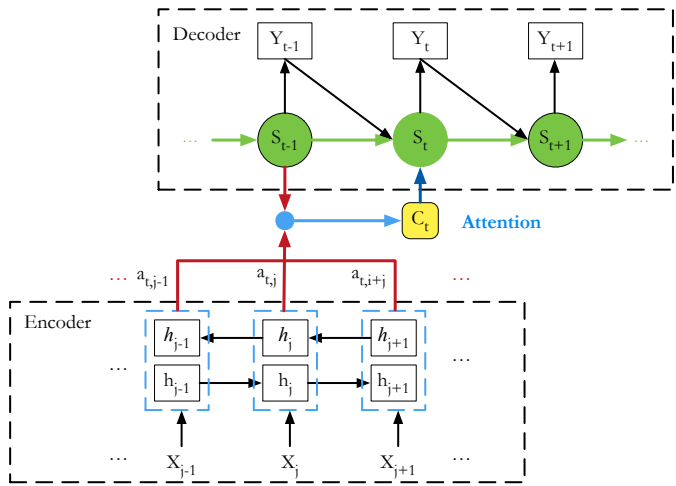} \vspace{-20pt}
       \caption{Illustration for attention-based NMT.} \label{f:nmt} \vspace{-10pt}
  \end{center}
\end{figure}

\section{Background}\label{back}
Our work is built upon the attention-based NMT~\cite{cho}, which takes a sequence of vector representations of the source sentence generated by a RNN or bi-directional RNN as input, and then jointly learns to align and translate by reading the vector representations during translation with a RNN decoder. Therefore, we take an overview of the attention-based NMT in this section before detail the NMT$_\textsf{IA}$ in next section.

\subsection{Attention-based Neural Machine Translation}\label{attention}
Figure~\ref{f:nmt} shows the framework of attention-based NMT.  Formally, given an input source sequence $\mathbf{x} \hspace{-3pt}=\hspace{-3pt} \{x_1,x_2,\cdots, x_N\}$ and the previously generated target sequence $\mathbf{y_{<t}} \hspace{-3pt}=\hspace{-3pt} \{y_1,y_2,\cdots, y_{t-1}\}$, the probability of the next target word $y_t$ is
\begin{eqnarray}
p(y_t|\mathbf{y_{<t}}, \mathbf{x}) = softmax(f(\mathbf{c}_t, y_{t-1}, \mathbf{s}_t)) \label{predict}
\end{eqnarray}
where $f(\cdot)$ is a non-linear function, and $\mathbf{s}_t$ is the state of decoder RNN at time step $t$ which is calculated as
\begin{eqnarray}
\mathbf{s}_t = g(\mathbf{s}_{t-1}, y_{t-1}, \mathbf{c}_t)
\end{eqnarray}
where $g(\cdot)$ can be any activation function, here we adopt a more sophisticated dynamic operator as in Gated Recurrent Unit (GRU)~\cite{ChoEMNLP}. In the remainder of the paper, we will also use GRU to stand for the operator. And $\mathbf{c}_t$ is a distinct source representation for time $t$, calculated as a weighted sum of the source annotations:
\begin{eqnarray}
\mathbf{c}_t = \sum_{j=1}^{N}{a_{t,j} \mathbf{h}_j}
\end{eqnarray}
Formally, $\mathbf{h}_j=[\overrightarrow{\mathbf{h}_j}^T, \overleftarrow{\mathbf{h}_j}^T]^T$ is the annotation of $x_j$, which is computed by a bi-directional RNN~\cite{schuster1997bidirectional} with GRU and contains information about the whole input sequence with a strong focus on the parts surrounding $x_j$. And its weight $a_{t,j}$ is computed by
\begin{eqnarray}
a_{t,j}=\frac{exp(e_{t,j})}{\sum_{k=1}^{N}{exp{(e_{t,k})}}}
\end{eqnarray}
where $e_{t,j}=\mathbf{v}_a^Ttanh(\mathbf{W}_a \mathbf{s}_{t-1} + \mathbf{U}_a  \mathbf{h}_j)$ scores how well $\mathbf{s}_{t-1}$ and $\mathbf{h}_j$ match. This is called automatic alignment~\cite{cho} or attention model~\cite{luongEMNLP2015}, but it is essentially reading with content-based addressing defined in~\cite{ntmgraves2014neural}. With the attention model, it releases the need to summarize the entire sentence with a single fixed-length vector~\cite{googleS2S,ChoEMNLP}. Instead, it lets the decoding network focus on one particular segment in source sentence at one moment, and therefore better resolution.

\begin{figure}[t!]
\begin{center}
      \includegraphics[width=0.75\textwidth]{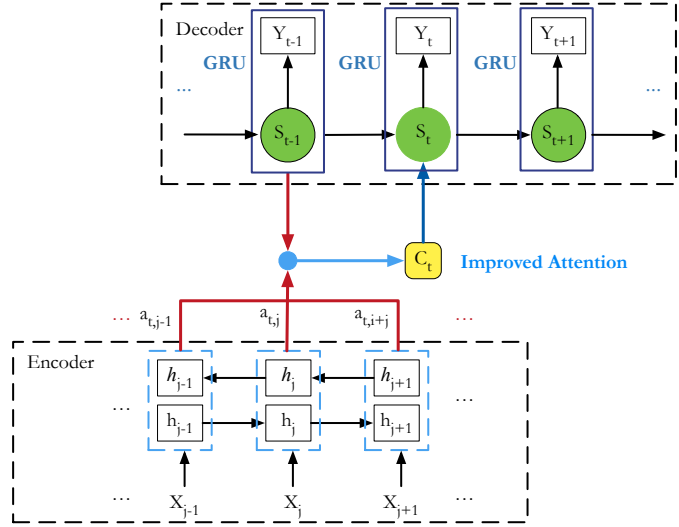}  \vspace{-20pt}
       \caption{Illustration for improved attention model of NMT.} \label{f:improve_nmt} \vspace{-10pt}
  \end{center}
\end{figure}

\subsection{Improved Attention Model}\label{improved-attention}
The alignment model $a_{t,j}$ scores how well the output at position $t$ matches the inputs around position $j$ based on $\mathbf{s}_{t-1}$ and $\mathbf{h}_j$. Intuitively, it should be beneficial to directly exploit the information of $y_{t-1}$ when reading from the representation of source sentence, which is not implemented in the original attention-based NMT~\cite{cho}. As illustrated in Figure~\ref{f:improve_nmt}, we add this implementation into the attention model, inspired by the latest implementation of attention-based NMT\footnote{https://github.com/nyu-dl/dl4mt-tutorial/tree/master/session2}. This kind of attention model can find a more  effective alignment path by using both previous hidden state $\mathbf{s}_{t-1}$ and the previous context word $y_{t-1}$. Then, the calculation of $e(t,j)$ becomes
\begin{eqnarray}
e_{t,j}=\mathbf{v}_a^Ttanh(\mathbf{W}_a \tilde{s}_{t-1} + \mathbf{U}_a  \mathbf{h}_j)
\end{eqnarray}
where $\tilde{\mathbf{s}}_{t-1}=\mathbf{GRU}(\mathbf{s}_{t-1}, \mathbf{e}_{y_{t-1}})$ is an intermediate state tailored for reading from the representation of source sentence with the information of $y_{t-1}$ (its word embedding being $\mathbf{e}_{y_{t-1}}$) added. And the calculation of update-state $\mathbf{s_{t}}$ becomes
\begin{eqnarray}
\mathbf{s}_t = \mathbf{GRU}(\tilde{\mathbf{s}}_{t-1}, \mathbf{c}_t)
\end{eqnarray}


\section{Interactive Attention}\label{interactive}
\begin{figure}[t!]
\begin{center}
      \includegraphics[width=0.85\textwidth]{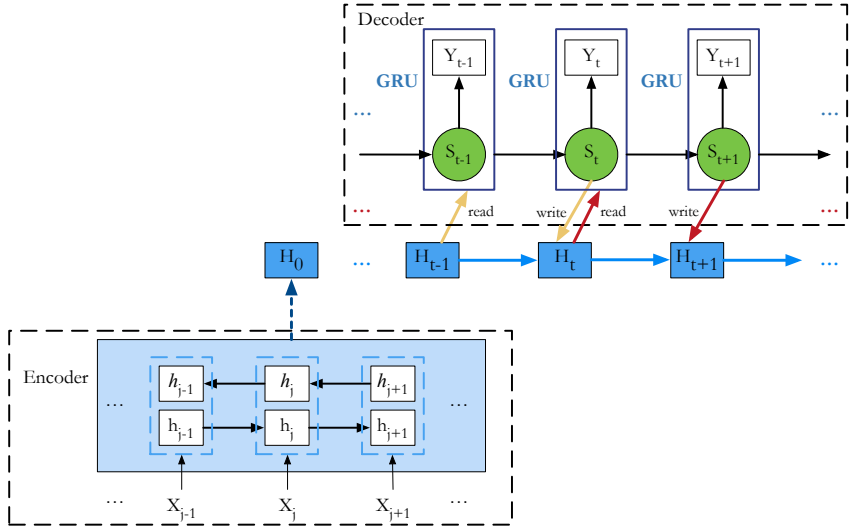}  \vspace{-20pt}
       \caption{Illustration for the NMT$_\textsf{IA}$. The yellow and red arrows indicate two rounds of interactive read-write operations.} \label{f:inter}
  \end{center}
\end{figure}

In this section, we will elaborate on the proposed \textsc{Interactive Attention}-based NMT, called NMT$_\textsf{IA}$. Figure~\ref{f:inter} shows the framework of NMT$_\textsf{IA}$ with two rounds of interactive read-write operations (indicated by the yellow and red arrows respectively), which adopts the same prediction model (Eq.~\ref{predict}) with improved attention-based NMT. With annotations $\mathbf{H} \hspace{-3pt}=\hspace{-3pt} \{\mathbf{h}_1, \mathbf{h}_2, \dots, \mathbf{h}_N\}$ of the source sentence $\mathbf{x} \hspace{-3pt}=\hspace{-3pt} \{x_1,x_2,\cdots, x_N\}$, we take $\mathbf{H}$ as a memory, which contains $N$ cells with the $j$th cell being $\mathbf{h}_j$.  As illustrated in Figure~\ref{f:inter}, \textsc{Interactive Attention} in NMT$_\textsf{IA}$ contains two key parts at each time step $t$: 1) attentive reading from $\mathbf{H}$, and 2) attentive writing to $\mathbf{H}$. Since the content in $\mathbf{H}$ changes with time, we will add time stamp on $\mathbf{H}$ (hence $\mathbf{H}^{(t)}$) and its cells (hence $\mathbf{h}_j^{(t)}$).

At time $t$, the state $\mathbf{s}_{t-1}$ first meets the prediction $y_{t-1}$ to form an ``intermediate" state $\tilde{\mathbf{s}}_{t-1}$, which can be calculated as follows
\begin{eqnarray}
\tilde{\mathbf{s}}_{t-1}=\mathbf{GRU}(\mathbf{s}_{t-1}, \mathbf{e}_{y_{t-1}})
\end{eqnarray}
where $\mathbf{e}_{y_{t-1}}$ is the word-embedding associated with the previous prediction word $y_{t-1}$. This ``intermediate" state $\tilde{\mathbf{s}}_{t-1}$ is used to read the source memory $\mathbf{H}^{(t-1)}$
\begin{eqnarray}
\mathbf{c}_{t}=\mathbf{Read}(\tilde{\mathbf{s}}_{t-1}, \mathbf{H}^{(t-1)}) \label{read}
\end{eqnarray}
After that, $\tilde{\mathbf{s}}_{t-1}$ is combined with $\mathbf{c}_{t}$ to update the new state
\begin{eqnarray}
\mathbf{s}_{t}=\mathbf{GRU}(\tilde{\mathbf{s}}_{t-1}, \mathbf{c}_{t})
\end{eqnarray}
Finally, the new state $\mathbf{s}_{t}$ is used to update the source memory by writing to it to finish the interaction in a round of state-update
\begin{eqnarray}
\mathbf{H}^{(t)}=\mathbf{Write}(\mathbf{s}_{t}, \mathbf{H}^{(t-1)}) \label {write}
\end{eqnarray}
The details of $\mathbf{Read}$ and $\mathbf{Write}$ in Eq.~\ref{read} and~\ref{write} will be described later in next section.

From the whole framework of NMT$_\textsf{IA}$, we can see that the new attention mechanism can timely update the representation of source sentence along with the update-chain of the decoder RNN state. This may let the decoder keep track of the attention history during translation. Clearly, \textsc{Interactive Attention} can subsume the coverage models in~\cite{Tu2016} as special cases while conceptually simpler. Moreover, with the attentive writing, \textsc{Interactive Attention} potentially can modify and add more on the source representation than just history of attention, and is therefore a more powerful model for machine translation, as empirically verified in Section 4.

\subsection{Read and Write of Interactive Attention} \label{rw_ia}
\paragraph{Attentive Read} Formally, $\mathbf{H}^{(t')} \in \mathbb{R}^{n \times m}$ is the memory in time $t'$ after the decoder RNN state update, where $n$ is the number of memory cells and $m$ is the dimension of vector in each cell. Before the state $\mathbf{s}$ update at time $t$, the output of reading $\mathbf{c}_t$ is given by
\begin{eqnarray}
\mathbf{c}_t=\sum_{j=1}^{n}\mathbf{w}_t^R(j)\mathbf{h}^{(t-1)}_j
\end{eqnarray}
where $\mathbf{w}_t^R \in \mathbb{R}^{n}$ specifies the normalized weights assigned to the cells in $\mathbf{H}^{(t-1)}$. We can use content-based addressing to determine $\mathbf{w}_t^R$ as described in~\cite{ntmgraves2014neural} or (quite similarly) use the reading mechanism such as the attention model in Section~\ref{back}. In this paper, we adopt the latter one.\footnote{ \newcite{WangLLL16} verified the former one for the read operation on the external memory.}

\paragraph{Attentive Write} Inspired by the writing operation of neural turing machines~\cite{ntmgraves2014neural}, we define two types of operation on writing to the memory: \textsc{Forget} and \textsc{Update}.
\textsc{Forget} is similar to the forget gate in GRU, which determines the content to be removed from memory cells. More specifically, the vector $\mathbf{F}_t \in \mathbb{R}^{m}$ specifies the values to be forgotten or removed on each dimension in memory cells, which is then assigned to each cell through normalized weights $\mathbf{w}_t^W$. Formally, the memory (``intermediate") after \textsc{Forget} operation is given by
\begin{eqnarray}
\tilde{\mathbf{h}}^{(t)}_i =\mathbf{h}^{(t-1)}_i(1-\mathbf{w}_t^W(i)\cdot \mathbf{F}_t),  \hspace{20pt} i = 1, 2, \cdots, n
\end{eqnarray}
where
\begin{itemize}
\item $\mathbf{F}_t=\sigma(\mathbf{W}_F, s_t)$ is parameterized with $\mathbf{W}_F \in \mathbb{R}^{m \times m}$, and $\sigma$ stands for the $Sigmoid$ activation function;
\item $\mathbf{w}_t^W \in \mathbb{R}^{n}$ specifies the normalized weights assigned to the cells in $\mathbf{H}^{(t)}$, and $\mathbf{w}_t^W(i)$ specifies the weight associated with the $i$th cell in the same parametric form as $\mathbf{w}_t^R$.
\end{itemize}

\textsc{Update} is similar to the update gate in GRU, deciding how much current information should be written to the memory as the added content
\begin{eqnarray}
\mathbf{h}^{(t)}_i=\tilde{\mathbf{h}}^{(t)}_i + \mathbf{w}_t^W(i)\cdot \mathbf{U}_t,  \hspace{20pt} i = 1, 2, \cdots, n
\end{eqnarray}
where $\mathbf{U}_t=\sigma(\mathbf{W}_U, \mathbf{s}_t)$ is parameterized with $\mathbf{W}_U \in \mathbb{R}^{m \times m}$, and $\mathbf{U}_t \in \mathbb{R}^{m}$. In our experiments, the weights for reading (i.e., $\mathbf{w}_t^R$) and writing (i.e., $\mathbf{w}_t^W$) at time $t$ are shared  when conducting interaction with the source memory.

\subsection{Optimization}
The parameters to be optimized include the embedding of words on source and target languages, the parameters for the encoder, the decoder and other operations of NMT$_\textsf{IA}$. The optimization is conducted via the standard back-propagation (BP) aiming to maximize the likelihood of the target sequence. In practice, we use the standard stochastic gradient descent (SGD) and mini-batch with learning rate controlled by AdaDelta~\cite{adadelta}.

\section{Experiments}\label{experiments}
We report our empirical study of NMT$_\textsf{IA}$ on Chinese-to-English translation task in this section. The experiments are designed to answer the following questions:
\begin{itemize}
\item Can NMT$_\textsf{IA}$ achieve significant improvements over the conventional attention-based NMT?
\item Can NMT$_\textsf{IA}$ outperform the attention-based NMT with coverage model~\cite{Tu2016}?
\end{itemize}

\subsection{Data and Metric}
Our training data consist of 1.25M sentence pairs extracted from LDC corpora\footnote{The corpora include LDC2002E18, LDC2003E07, LDC2003E14, Hansards portion of LDC2004T07, LDC2004T08 and LDC2005T06.}, with 27.9M Chinese words and 34.5M English words respectively. We choose NIST 2002 (MT02) dataset as our development set, which is used to monitor the training process and decide the early stop condition. And the NIST 2003 (MT03), 2004 (MT04), 2005 (MT05), 2006 (MT06) datasets are used as our test sets. The numbers of sentences in NIST MT02, MT03, MT04, MT05 and MT06 are 878, 919, 1788, 1082, and 1664 respectively. We use the case-insensitive 4-gram NIST BLEU\footnote{\url{ftp://jaguar.ncsl.nist.gov/mt/resources/mteval-v11b.pl}} as our evaluation metric, with statistical significance test (\emph{sign-test}~\cite{collins2005clause}) between the proposed models and the baselines.


\subsection{Training Details}
In training the neural networks, we limit the source and target vocabulary to the most frequent 30K words for both Chinese and English, covering approximately 97.7\% and 99.3\% of two corpus respectively. All the out-of-vocabulary words are mapped to a special token \texttt{\small UNK}. We initialize the recurrent weight matrices as random orthogonal matrices. All the bias vectors are initialized to zero. For other parameters, we initialize them by sampling each element from the Gaussian distribution of mean 0 and variance $0.01^2$. The parameters are updated by SGD and mini-batch (size 80) with learning rate controlled by AdaDelta~\cite{adadelta} ($\epsilon=1e^{-6}$ and $\rho=0.95$). We train the NMT systems with the sentences of length up to 50 words in training data, and set the dimension of word embedding to 620 and the size of the hidden layer to 1000, following the settings in~\cite{cho}. We also use dropout for our baseline NMT systems and NMT$_\textsf{IA}$ to avoid over-fitting~\cite{hinton2012improving}. In our experiments, dropout was applied on the output layer with dropout rate setting to 0.5.

Inspired by the effort on easing the training of very deep architectures~\cite{hinton2006reducing}, we use a simple pre-training strategy to train our NMT$_\textsf{IA}$. First we train a regular attention-based NMT model~\cite{cho}. Then we use the trained NMT model to initialize the parameters of NMT$_\textsf{IA}$ except for those related to the operations of \textsc{interactive attention}. After that, we fine-tune all the parameters of NMT$_\textsf{IA}$.


\begin{table}[t!]
\begin{center}
\scalebox{0.95}{
\begin{tabular}{l|llll|l}
\hline
\textsc{Systems} & \textsc{MT03} & \textsc{MT04} & \textsc{MT05} & \textsc{MT06} & \textsc{Average}  \\
\hline\hline
Moses                 				& 31.61     & 33.48     & 30.75     & 31.07  & 31.73 \\
Groundhog  	  				& 30.96     & 33.09     & 30.61     & 31.12  & 31.45 \\
RNNsearch$^{\star}$  	  		& 33.42     & 36.04     & 33.60     & 32.24  & 33.83 \\
NMT$_{\textsf{IA}}$                           & 35.09*    & 37.73*    & 35.53*    & 34.32*  & 35.67 \\
\hline
\end{tabular}
}
\end{center}
\caption{\label{t:main-result} BLEU-4 scores (\%) of the phrase-based SMT system (Moses), NMT baselines: Groundhog and RNNsearch$^{\star}$ (our implementation of improved attention model as described in Section~\ref{improved-attention}),  and our \textsc{Interactive Attention} model (NMT$_{\textsf{IA}}$). The ``*" indicates that the results are significantly (p$<$0.01) better than those of all the baseline systems.
}
\end{table}

\subsection{Comparison Systems}
We compare our NMT$_{\textsf{IA}}$ with four systems:
\begin{itemize}
\item {\bf Moses}~\cite{koehn2007}: an open source phrase-based translation system\footnote{http://www.statmt.org/moses/} with default configuration. The word alignments are obtained with GIZA++~\cite{och2003systematic} on the training corpora in both directions, using the ``grow-diag-final-and" balance strategy~\cite{koehn2003statistical}. The 4-gram language model with modified Kneser-Ney
smoothing is trained on the target portion of training data with the SRILM toolkit~\cite{stolcke2002srilm},
\item {\bf Groundhog}: an open source NMT system\footnote{https://github.com/lisa-groundhog/GroundHog} implemented with the conventional attention model~\cite{cho}.
\item {\bf RNNsearch$^{\star}$}: our in-house implementation of NMT system with the improved conventional attention model as described in Section~\ref{improved-attention}.
\item {\bf Coverage Model}: state-of-the-art variants of attention-based NMT model~\cite{Tu2016} which improve the attention mechanism through modeling a soft coverage on the source representation by maintain a coverage vector to keep track of the attention history during translation.
\end{itemize}

\subsection{Main Results}
The main results of different models are given in Table~\ref{t:main-result}. Before proceeding to more detailed comparisons, we first observe that
\begin{itemize}
\item RNNsearch$^{\star}$ outperforms Groundhog, which is implemented with the conventional attention model as described in Section~\ref{attention}, by 2.38 BLEU points averagely on four test sets;
\item RNNsearch$^{\star}$ only exploit sentences of length up to 50 words with 30K vocabulary, but can achieve averagely 2.10 BLEU points higher than the open source phrase-based system Moses, which is trained with full training data.
\end{itemize}

Clearly from Table~\ref{t:main-result}, NMT$_{\textsf{IA}}$ can achieve significant improvements over RNNsearch$^{\star}$ by 1.84 BLEU points averagely on four test sets. We conjecture it is because our \textsc{Interactive Attention} mechanism can keep track of the interaction history between the decoder and the representation of source sentence during translation, which may be helpful for the decoder to automatically distinguish which parts have been translated and which parts are under-translated.

\begin{table}[t!]
\begin{center}
\scalebox{0.95}{
\begin{tabular}{l|llll|l}
\hline
\textsc{Systems} & \textsc{MT03} & \textsc{MT04} & \textsc{MT05} & \textsc{MT06} & \textsc{Average}  \\
\hline\hline
RNNsearch$^{\star}$-80  	  	            	  & 33.34       & 37.10         & 33.38         & 33.70    & 34.38 \\
NN-Cover-80 	    	   		   	  	  & 33.69       & 38.05         & 35.01         & 34.83    & 35.40 \\
NMT$_{\textsf{IA}}$-80             		          & 35.69*+    & 39.24*+     & 35.74*+      & 35.10*  & 36.44 \\
\hline
\end{tabular}
}
\end{center}
\caption{\label{t:result-cpm-coverage} BLEU-4 scores (\%) of the conventional attention-based model (RNNsearch$^{\star}$-80), the neural network based coverage model (NN-Cover-80) \protect\cite{Tu2016} and our \textsc{Interactive Attention} model (NMT$_{\textsf{IA}}$-80). ``-80" means the models are trained with the sentences of length up to 80 words, which is consistent with the setting in \protect\cite{Tu2016}. The ``*" and ``+" denote that the results are significantly (p$<$0.01) better than those of RNNsearch$^{\star}$-80 and NN-Cover-80 respectively.
}
\end{table}

\subsection{\textsc{Interactive Attention} Vs. Coverage Model}
\newcite{Tu2016} proposed two coverage models to let the NMT system to consider more about untranslated source words. Basically, they maintain a coverage vector for each hidden state for source to keep track of the attention history and feed the coverage vector to the attention model to help adjust future attention. Although we do not maintain a coverage vector, our \textsc{Interactive Attention} can potentially do similar things, therefore subsuming coverage models as special cases. We hence compare our \textsc{Interactive Attention} model with the coverage model in~\cite{Tu2016}. There are two coverage models proposed in~\cite{Tu2016}, including linguistic coverage model and neural network based coverage model (NN-Cover). Since the neural network based coverage model generally yields better results, we mainly compare with the neural network based coverage model. Although the coverage models are originally implemented on Groundhog in~\cite{Tu2016}, they can be easily adapted to the ``RNNsearch$^{\star}$". Following the setting in~\cite{Tu2016}, we conduct the comparison with the training sentences of length up to 80 words. Clearly from Table~\ref{t:result-cpm-coverage}, our NMT$_{\textsf{IA}}$-80 outperforms the NN-Cover-80 by +1.04 BLEU scores averagely on four test sets.

\begin{figure}[t!]
\begin{center}
      \includegraphics[width=0.6\textwidth]{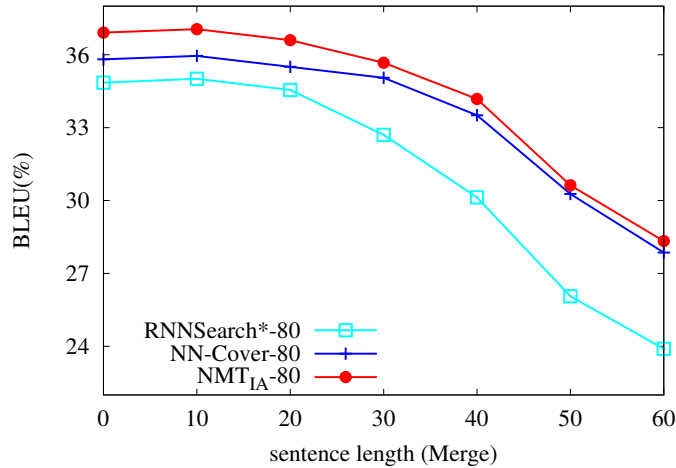} \vspace{-10pt}
       \caption{The BLEU-4 scores (\%) of generated translations on the merged four test sets with respect to the lengths of source sentences. The numbers on X-axis of the figure stand for sentences \emph{longer than} the corresponding length, e.g., $40$ for source sentences with $>40$ words.} \label{f:res_length} \vspace{-10pt}
  \end{center}
\end{figure}

A more detailed comparison between conventional attention model (RNNsearch$^{\star}$-80), neural network based coverage model  (NN-Cover-80)~\cite{Tu2016} and NMT$_{\textsf{IA}}$-80 suggests that our NMT$_{\textsf{IA}}$-80 is quite consistent on outperforming the conventional attention model and the coverage model. Figure~\ref{f:res_length} shows the BLEU scores of generated translations on the test sets with respect to the length of the source sentences. In particular, we test the BLEU scores on sentences longer than \{0, 10, 20, 30, 40, 50, 60\} in the merged test set of  MT03, MT04, MT05 and MT06. Clearly, on sentences with different length, NMT$_{\textsf{IA}}$-80 always yields consistently higher BLEU scores than the conventional attention-based NMT and the enhanced version with the neural network based coverage model. We conjecture that with the attentive writing (described in Section~\ref{rw_ia}), \textsc{Interactive Attention} potentially can modify and add more on the source representation than just history of attention, and is therefore a more powerful model for machine translation.

We also provide some actual translation examples (see Appendix) to show that our \textsc{Interactive Attention} can get better performance then baselines, especially on solving under-translation problem. We think the  interactive mechanism of NMT$_\textsf{IA}$ is helpful for the decoder to automatically distinguish which parts have been translated and which parts are under-translated.

\section{Related Work}\label{related}
Our work is related to recent works that focus on improving attention models~\cite{luongEMNLP2015,cohnEtAl2016,FengLLZ16}. \newcite{luongEMNLP2015} proposed to use global and local attention models to improve translation performance. They use a global one to attend to all source words and a local one to look at a subset of source words at a time. \newcite{cohnEtAl2016} extended the attention-based NMT to include structural biases from word-based alignment models, which achieved improvements across several language pairs. \newcite{FengLLZ16} added implicit distortion and fertility models to
attention-based NMT to achieve translation improvements. These works are different with our \textsc{Interactive Attention} approach, as we use a rather generic attentive reading while at the same time performing attentive writing.

Our work is inspired by recent efforts on attaching an external memory to neural networks, such as neural turing machines~\cite{ntmgraves2014neural}, memory networks~\cite{WestonCB14,MengLTLL15} and exploiting an external memory~\cite{TangMLLY16,WangLLL16} during translation. \newcite{TangMLLY16} exploited a phrase memory for NMT, which stores phrase pairs in symbolic form. They let the decoder utilize a mixture of word-generating and phrase-generating component, to generate a sequence of multiple words all at once. \newcite{WangLLL16} extended the NMT decoder by maintaining an external memory, which is operated by reading and writing operations of neural turing machines~\cite{ntmgraves2014neural}, while keeping a read-only copy of the original source annotations along side the ``read-write" memory. These powerful extensions have been verified on Chinese-English translation tasks. Our \textsc{Interactive Attention} is different from previous works. We take the annotations of source sentence as a memory instead of using an external memory, and we design a mechanism to directly read from and write to it during translation. Therefore, the original source annotations are not accessible in later steps. More specially, our model inherited the notation and some simple operations for writing from~\cite{ntmgraves2014neural}, while NMT$_\textsf{IA}$ extends it to ``unbounded" memory for representing the source. In addition, although the read-write operations in \textsc{Interactive Attention} are not exactly the same with those in~\cite{ntmgraves2014neural,WangLLL16}, our model can also achieve good performance.

\section{Conclusion}\label{conclusion}
We propose a simple yet effective~\textsc{Interactive Attention} approach, which models the interaction between the decoder and the representation of source sentence during translation by using reading and writing operations. Our empirical study on Chinese-English translation shows that \textsc{Interactive Attention} can significantly improve the performance of NMT.

\section*{Acknowledgements}
Liu is partially supported by the Science Foundation Ireland (Grant 13/RC/2106) as part of the ADAPT Centre at Dublin City University. We sincerely thank the anonymous reviewers for their thorough reviewing and valuable suggestions.

\bibliography{coling2016}
\bibliographystyle{acl}

\newpage

\section*{APPENDIX:  Actual Translation Examples}
In appendix we give some example translations from RNNsearch$^{\star}$-80, NN-Cover-80 and NMT$_{\textsf{IA}}$-80, and compare them against the reference. We highlight some correct translation segments (or under-translated by baseline systems) in blue color and wrong ones in red color.

\subsection*{Example Translations}
\begin{figure}[h!]
\begin{center}
 \includegraphics[width=0.8\textwidth]{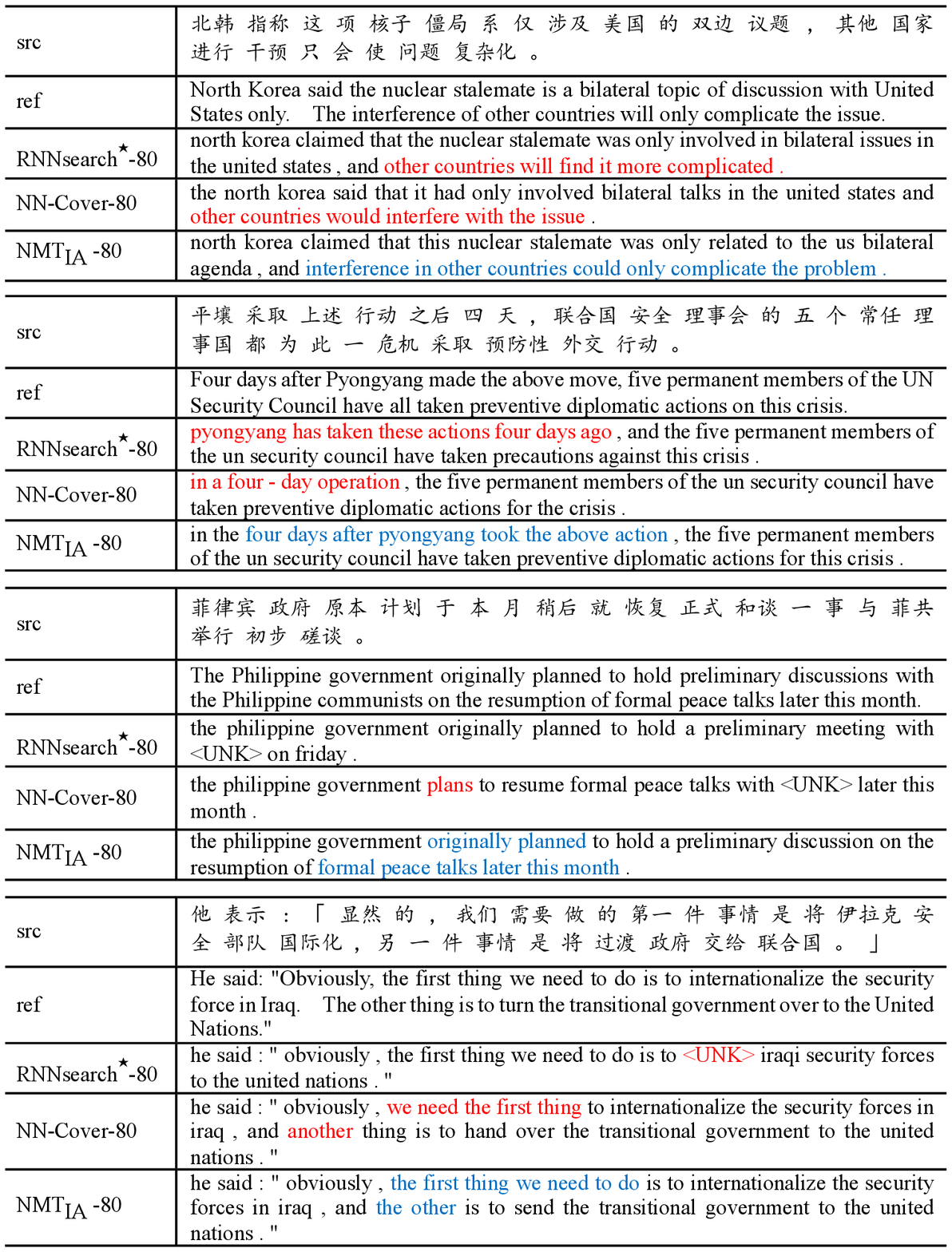} \label{f:example}
  \end{center}
\end{figure}
\newpage

\end{document}